\documentclass[letterpaper, 10 pt, conference]{ieeeconf}

\IEEEoverridecommandlockouts %
\usepackage{times}

\usepackage{multicol}
\usepackage{booktabs}
\usepackage{graphicx}
\usepackage{float}
\usepackage{amsmath, amssymb}
\usepackage{amsfonts}
\usepackage{bm}
\usepackage{algpseudocode}
\usepackage{algorithm}
\usepackage{multirow}
\usepackage{overpic}
\usepackage{wrapfig}

\usepackage[normalem]{ulem} %

\usepackage[noadjust]{cite}

\usepackage{color}
\definecolor{international_orange}{RGB}{240, 74, 0}
\definecolor{citecolor}{RGB}{240, 74, 0}
\usepackage[pagebackref=true,breaklinks=true,urlcolor=citecolor,colorlinks,citecolor=citecolor,bookmarks=false,linkcolor=citecolor]{hyperref}

\newcommand{\xv}{\vec{x}} %
\newcommand{\gv}{\vec{g}} %

\newcommand{\glatent}{\xv}

\newcommand{\glatentprimehat}{\hat{\xv}^{\prime}}

\renewcommand{\vec}{\mathbf}

\newcommand{\tpar}[1]{\vspace{2pt}\noindent\textbf{#1}}

\newcommand{\pca}{DOOM}
\newcommand{\lsa}{LOOM}
\newcommand{\transa}{Baseline}

\usepackage{subcaption}
\usepackage[skip=0pt,font=small,labelfont=bf]{caption}
\usepackage{amsmath}

\newcommand{\shrinka}

\definecolor{revision_blue}{RGB}{0, 0, 200}

\title{Out of Sight, Still in Mind: Reasoning and Planning about Unobserved Objects with Video Tracking Enabled Memory Models}

\begin{document}
\author{%
  Yixuan Huang$^{1}$,
  Jialin Yuan$^{2}$, 
  Chanho Kim$^{2}$, 
  Pupul Pradhan$^{1}$,
  Bryan Chen$^{2}$, 
  Li Fuxin$^{2}$, 
  and Tucker Hermans$^{1, 3}$ 
  \thanks{$^{1}$%
    Robotics Center and Kahlert School of Computing,
    University of Utah,
    Salt Lake City, UT 84112, USA.
    $^{2}$ Oregon State University. 
    $^{3}$%
    NVIDIA; Seattle, WA, USA.
    \protect\url{yixuan.huang@utah.edu}
  }%
}

\input{overview_fig}
\maketitle

\begin{abstract}
Robots need to have a memory of previously observed, but currently occluded objects to work reliably in realistic environments.
We investigate the problem of encoding object-oriented memory into a multi-object manipulation reasoning and planning framework.
We propose \pca{} and \lsa{}, which leverage transformer relational dynamics to encode the history of trajectories given partial-view point clouds and an object discovery and tracking engine.
Our approaches can perform multiple challenging tasks including reasoning with occluded objects, novel objects appearance, and object reappearance.
Throughout our extensive simulation and real-world experiments, we find that our approaches perform well in terms of different numbers of objects and different numbers of distractor actions.
Furthermore, we show our approaches outperform an implicit memory baseline.
\end{abstract}

\section{Introduction}

For robots to assist humans in their daily lives in roles such as home assistants and caregivers for elders, they must be able to reason about objects not observed in their current perceptual data.
For example, if asked to retrieve an apple stored inside a cabinet, the robot should remember where it was placed and that it must first open the cabinet to grasp the object. Further if the robot picks up a box with objects in it, it should know that these objects will move with the box, while those sitting near the box will not. It is desirable for robots to maintain these models across multiple tasks in order to enable long-term autonomy.

The problem of maintaining a persistent model of a robot's world from partial observations has a rich history in robotics, primarily in the form of mapping~\cite{Moravec_1988,leonard-ijrr1992,Kuipers-2000-ssh,chown-1999-r-plan,cadena-tro2016-slam}. While some attempts have been made to incorporate object-level semantics into sensor-derived maps~\cite{Pillai-RSS-15,herbst-icra2014,faulhammer-ral2017}, these approaches primarily build immutable, monolithic models not suitable for complex manipulation planning such as multi-object rearrangement~\cite{weiyu-rss2023,zhu2020hierarchical, lin2022efficient, sharma2020relational,liu-icra2022-structformer,murali20206,paxton-corl2021-semantic-placement,Qureshi-RSS-21,Huang-icra2023-graph-relations,huang2023latent}.
Such reasoning can enable robots to predict preconditions for actions~\cite{sharma2020relational}, achieve language goals~\cite{liu-icra2022-structformer}, and achieve logical goal relations~\cite{huang2023latent, paxton-corl2021-semantic-placement}. Thus a problem arises in how to build a consistent estimate of the world state amenable to contemporary, learning-based approaches to manipulation planning.

The standard learning-based approach to persistent memory treats the problem as a sequence prediction task, typically using autoregressive recurrent or transformer-based neural networks~\cite{bonatti2022pact, wani2020multion, cartillier2021semantic, ebert2017self, kim2022memory}. While these models have shown some interesting results, we hypothesize that they will have difficulty maintaining long-term memory of objects that remain occluded through many updates to a recurrent memory. One approach to overcome this issue would be to maintain the entire history of robot observations in order to use as input to a model (e.g. transformer) that could selectively attend to the relevant bits of input to determine a coherent state estimate. This is obviously untenable for long-horizon tasks as the input would grow linearly with time, inevitably becoming too large to efficiently manage.

\begin{figure*}
    \centering
\includegraphics[width=1.9\columnwidth]
    {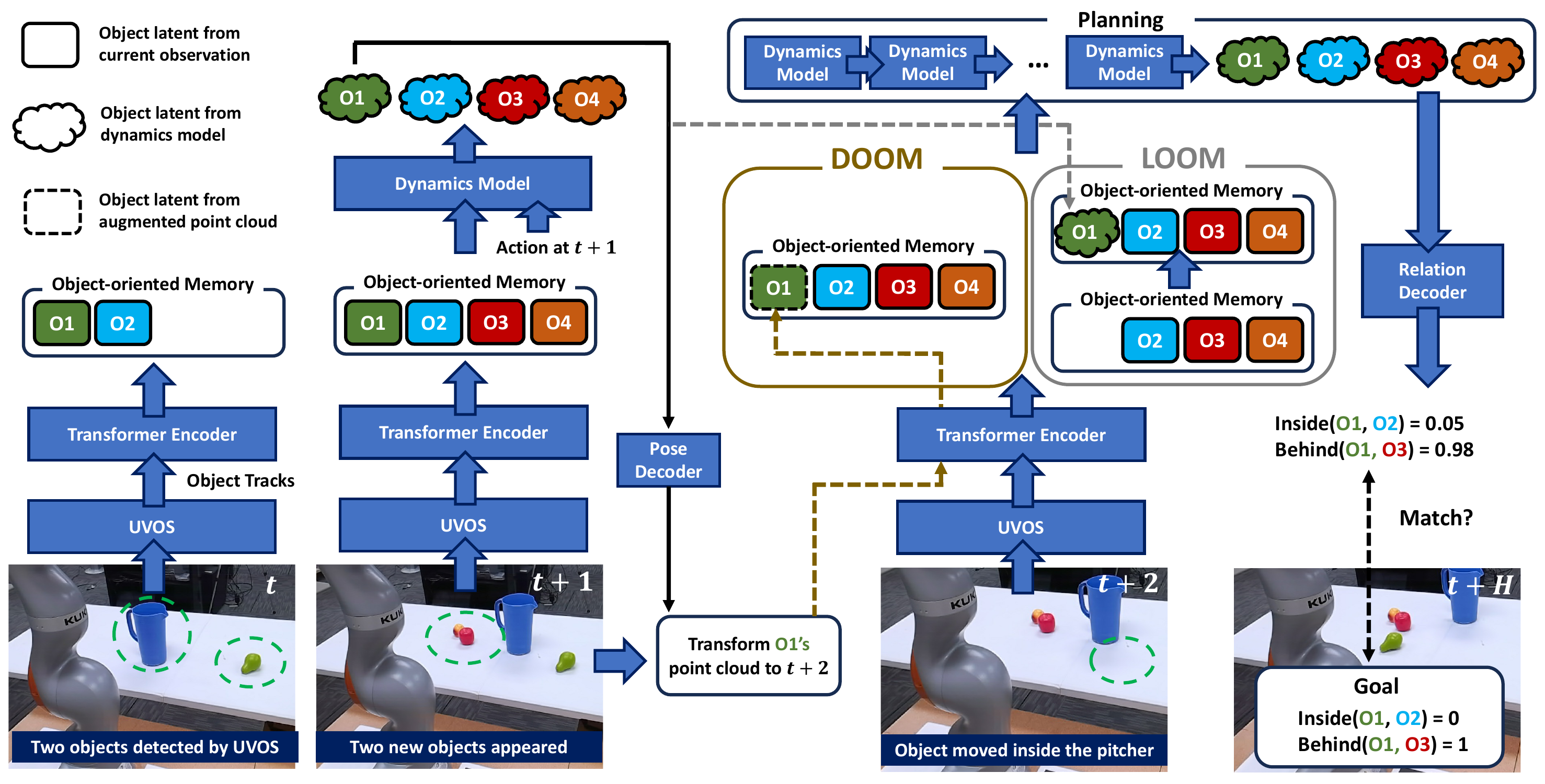}
    \caption{Overview of our approaches. As the robot takes action over time, some objects may disappear and reappear, and some objects may newly appear in the scene. In this paper, we propose two types of object-oriented memory, called \pca{} and \lsa{}, that enable the robot to plan with occluded and newly appeared objects. \pca{} and \lsa{} utilize a UVOS algorithm to keep track of the current object list and update object memory slots based on the occlusion status of each object accordingly. (Best viewed in color.)}
    \label{fig:network_structure}\vspace{-20pt}
\end{figure*}

We instead argue for an explicit way to encode the state of all currently and previously observed objects in a consistent manner. In recent years, there have been promising solutions for the  unsupervised video object segmentation (UVOS)~\cite{Caelles_arXiv_2019}
task where the algorithm simultaneously discovers previously unknown objects and tracks them through time~\cite{luiten2020unovost, Lin_2021_ICCV, Zhou_2021_CVPR, yuan2023maximal}. Especially interesting for our sake, such algorithms allow for tracking under heavy and long-term complete occlusions. This opens up an opportunity to use a UVOS algorithm in robot manipulation tasks where we may not have segmentation labels for objects of interest. 

In this paper we advocate for using a UVOS algorithm to explicitly manage our object-oriented memory. We hypothesize that explicit encoding of objects into the memory will be more robust than implicit autoregressive models in managing long-term history and successfully performing downstream planning. The representation should provide sufficient information to enable the prediction of inter-object and object-environment relations, while also having the ability to predict the effects of the robot's actions on all objects in memory. 

Specifically, 
we examine incorporating an explicit UVOS-based memory model into the framework of Huang et al.~\cite{Huang-icra2023-graph-relations,huang2023latent}, which was an effective framework to learn relational dynamics across varying object and environments. %
Key to its success is the ability to encode a variable number of objects for a given observation using a graph neural net~\cite{Huang-icra2023-graph-relations} or transformer-based encoder~\cite{huang2023latent}.
Prediction of relations enables the model's use in logic-based task planning~\cite{garrett2020pddlstream}, where relations have proved an effective means of communication between robots and humans~\cite{paxton-corl2021-semantic-placement,liu-icra2022-structformer,zhu2020hierarchical,li2020towards, sharma2020relational,Huang-icra2023-graph-relations}.
However, the existing framework assumes all relevant objects to be observable; and thus fails to successfully plan and execute in the scenarios shown in Fig.~1.

We propose two ways to integrate predictions from the video tracker of~\cite{yuan2023maximal} with the relational dynamics prediction model of~\cite{huang2023latent}. Both approaches augment the current state estimate with information from currently unobserved objects for use in predicting inter-object relations and action effects.  One approach directly augments the latent space of the dynamics model by concatenating the previously predicted latent state tokens for unobserved objects with those currently observed. We term this \emph{Latent Occluded Object Memory} (\lsa{}). The second method, termed \emph{Direct Occluded Object Memory} (\pca{}), directly augments the input point cloud with the previously observed object point cloud transformed based on its previously predicted pose estimate.
Figure~\ref{fig:network_structure} illustrates the \lsa{} and \pca{} models.

To validate our proposed approaches, we test on varying tasks involving occluded objects, novel objects, and objects that reappear. We show some example tasks in Fig.~1.
We show that both forms of memory-based state augmentation can reliably plan actions to change inter-object relations involving currently unobserved objects, such as placing object \(A\) to the left of currently occluded object \(B\). Furthermore we compare our explicit memory approach to a baseline implicit memory model in a scenario specifically designed to test the models' long-term memory. Our experiments show that the implicit memory model fails to effectively reason about occluded objects, while our explicit, UVOS-based memory models continue to plan reliably.
Our work thus represents the first successful approach of using memory models for reasoning about and planning to inter-object goal relations for an a priori unknown, variable number of objects under severe occlusion.
For more real-world robot executions, additional results, and supplemental material including limitations and failure case analysis, see our website \url{https://sites.google.com/view/rdmemory}.

\section{Related work}\label{sec:related-work}
Reasoning about object permanence is an important capability for robot manipulation~\cite{xu2020learning, ebert2017self, du2022play, curtis2022long}.
Xu et al.~\cite{xu2020learning} and Ebert et al.~\cite{ebert2017self} are the first ones to propose deep learning models to reason about occluded objects and do downstream planning with the learned dynamics models. However, they assume goal images or goal configurations for planning which may not always be available from human operators. They additionally examine only planar pushing tasks with only a single moving object at each time.
Our work examines a much more diverse set of tasks and skills, while also requiring only logic-based goal representations.
Curtis et al.~\cite{curtis2022long} propose a system with modules that can estimate affordances and properties to perform multi-step manipulation tasks with unknown and occluded objects. However, they assume complete object shape and have many engineered modules including the affordance module.

Encoding memory for manipulation and navigation has received some attention~\cite{shafiullah2022clip, kim2022memory, jockel2009sparse, jockel2009robot, bessler2018knowledge}.
Shafiullah et al.~\cite{shafiullah2022clip} propose an implicit model for semantic navigation but this model cannot be directly applicable to manipulation tasks.
Kim et al.~\cite{kim2022memory} propose both RNNs and transformers to predict memory-based gaze via imitation learning. However, the work focuses on gaze prediction and shows no experiments with occluded objects or objects with novel appearances.

Reasoning about occluded objects with memory has received attention in the vision community in the context of visual tracking~\cite{Cai_2022_CVPR, Fang2017RecurrentAN,yuan2023maximal} and video object segmentation~\cite{cheng2022xmem, Oh_2019_ICCV}. However, these memory models have been studied in the context of object re-identification across occlusion only (i.e., appearance matching) and cannot be utilized for reasoning about how occluded objects may move based on a robot's actions. Our work builds on these approaches by combining them with a learned action effects model.

\begin{figure*}[t]
    \centering
    \includegraphics[width=2\columnwidth]{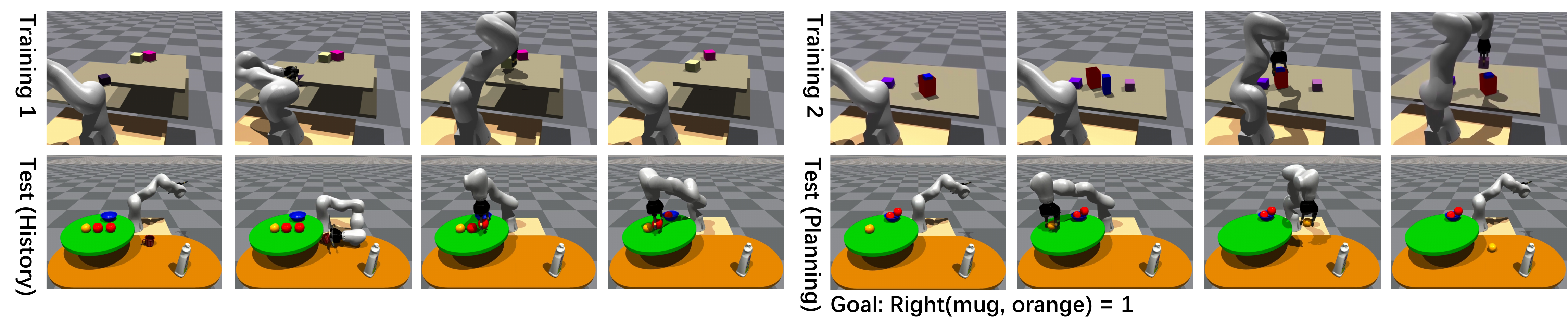}
    \vspace{-5pt}
    \caption{Two examples of our training dataset  (top row) and one testing example (bottom row). 
    We train with a maximum of 5 segments including objects and the environments. 
    The testing example has 8 segments with different shapes and a novel view point. In the history, the robot pushes the mug below the shelf then picks and places two apples inside the bowl. 
    During planning, the robot picks and places the orange to achieve the goal relation based on the current observation and history.
    Left/Right are defined from the robot's viewpoint.}
\label{fig:simulation_figure}\vspace{-20pt}
\end{figure*}

Other works have also investigated using graph neural networks~\cite{di2023one, Huang-icra2023-graph-relations, zhu2020hierarchical, chen2023predicting, kulshrestha2023structural}, transformers~\cite{liu-icra2022-structformer,yuan2022sornet, shridhar2023perceiver, zitkovich2023rt}, and diffusion models~\cite{weiyu-rss2023, simeonov2023shelving, chi2023diffusionpolicy} to reason about and manipulate multiple objects. However, none have examined explicit memory management.
Since multi-object manipulation usually requires a multi-step process, task and motion planning has shown promise in solving long-horizon, multi-object planning problem~\cite{kim2020learning, driess2020deep, garrett-icra2020, liang-icra2022, huang2023latent, yang2022sequence}. These planners either require all objects to be known a priori in order to perform tracking for belief space planning~\cite{garrett-icra2020} or ignore occlusion of unknown objects. %

\section{Approach}\label{sec:approach}%

We assume the robot perceives the world as a point cloud, $Z_{t}$, at each timestep \(t\). The robot then takes action $A_t$ and receives subsequent observation $Z_{t+1}$. 
At new observation $Z_{t+1}$, some objects may become occluded and other, new objects may appear. 
Based on the history of observations and actions $(Z_{0:t}$, $A_{0:t})$, we would like the robot to plan to achieve a goal, potentially involving previously observed, but currently occluded objects.

We define the goal as a logical conjunction of \(M\) desired object and environment relations,  $\gv = r_{1} \land r_{2} \land ... \land r_{M},  r_j \in \mathcal{R}$, where
$\gv$ denotes the goal conjunction, $r_j$ represents a goal relation, and $\mathcal{R}$ denotes all possible relations.
Our robot is given a set of \(L\) parametric action primitives $\mathcal{A} = \{A_1, \ldots, A_L\}$ where \(A_l\) defines a skill, which has associated continuous skill parameters $\theta_{l}$. For example, a push skill is defined with parameters encoding the push direction and length, or a pick-and-dump skill is defined with parameters encoding the grasp pose and dump pose.
The robot's planning task is defined as finding skills and its parameters $\tau = ((A_{0},\theta_{0}), \ldots, (A_{H-1},\theta_{H-1}))$ such that, when sequentially executed, transform the objects to satisfy all desired object and environment relations in the goal $\gv$.

To solve this problem we propose a novel memory-based neural network framework.
Instead of taking the entire history of observations as input, the model takes the current observation, \(Z_t\), current action \((A_t, \theta_t)\) and a compressed memory of the previous observations $Z_{0:t-1}$ and actions $A_{0:t-1}$, and predicts the resulting relations \(r_{t+1}^{\prime}\) and object poses \(p_{t+1}^{\prime}\). 
As shown in Fig. 1, this enables our framework to remember the pose of disappeared objects after several actions.  
By chaining together predictions, we can effectively perform multi-step planning. We propose two different implementations of this framework called \pca{} and \lsa{}, which respectively use a point cloud-based encoding and latent space encoding to represent the memory, \(Q\).
We show an overview of our approaches in Fig.~\ref{fig:network_structure}. We now explain the various components of this framework.

\tpar{Segmentation and Tracking}
At timestep $t$, we first perform UVOS using \cite{yuan2023maximal} and obtain \(N\) object and environment segments $O^{i}_t \subset Z_t, i = 1,2,...,N$. The UVOS approach checks for new objects by examining the objectness of object proposals on key frames, when the robot finishes the execution of a skill. Then object segments that can be matched with previously tracked objects are assigned consistent IDs, while those that do not match any previous object are assigned new IDs. For non-key frames, 
the algorithm tracks all segments with a space-time transformer model~\cite{Oh_2019_ICCV} where each object has its own memory. It stores prior appearances of all objects hence can easily re-identify previously observed and disappeared objects. To familiarize the UVOS model with the robot, it is fine-tuned  using annotations of the robot arm and objects from YCB-Video dataset \cite{xiang2017posecnn}. Nonetheless, it still demonstrates generalization to novel object types, such as the tall cups (Row.~1), apple and orange (Row.~2) shown in Fig.~1 that it was never trained on. %
For environment segments, we use RANSAC~\cite{fischler1981random} to find planar surfaces. %

\tpar{Pose and relation detection}
To jointly reason about the pose of each object and the relations between each pair of objects, we first process each segment $O^{i}_t$ with a point cloud encoder~\cite{wu2019pointconv} to get a feature vector $P^{i}_t = E_p(O^{i}_t)$. 
We use a learned positional embedding to encode the ID of each object as $I_i$. We randomly generate the ID for each object during training to improve generalization~\cite{cui2022positional}. 
The latent space at step \(t\) is denoted as $\glatent^{i}_{t} = E(P^{i}_t \oplus I_i)$, where \(E\) is a transformer encoding the interaction between different segments. 
To predict object pose and relations at the current step, we define three different decoders: the pose decoder $D_p$, the relation decoder $D_r$, and the environment identity decoder $D_e$.
The third decoder classifies whether a given segment is movable or not (e.g. is a table or shelf) as $\hat{e}^{i}_t = D_e(\glatent^{i}_{t})$. 
The associated predictions of the other decoders are $\hat{p}^{i}_t = D_p(\glatent^{i}_{t})$ and $\hat{r}^{ij}_t = D_r(\glatent^{i}_{t}, \glatent^{j}_{t})$. 

\begin{table*}[t]
\centering
\vspace{-5pt}
\caption{Comparison in terms of the F1 score on the relational predictions.}
\begin{tabular}{ p{1.8cm}p{0.8cm}p{0.8cm}p{0.8cm}p{0.8cm} p{0.8cm} p{1.8cm}p{0.8cm}p{0.8cm}p{0.8cm} p{0.8cm}}
 \hline
 Objects & 4 & 5 & 6 & 7 & all & Distractors & 1 & 2 & 3 & all\\
 \hline
 \pca{}   & \textbf{0.998} & 0.975 & \textbf{0.974} & \textbf{0.958} & \textbf{0.976} & \pca{} & \textbf{0.918} & \textbf{0.907} & \textbf{0.901} & \textbf{0.909}\\
 \lsa{}     & 0.994 & \textbf{0.978} & 0.972 & 0.951 & 0.974 & \lsa{}& 0.894 & 0.873 & 0.868 & 0.878\\
 \transa{}    & 0.938 & 0.786 & 0.765 & 0.702 & 0.798 & \transa{}    & 0.797 & 0.763 & 0.751 & 0.770\\
 \hline
\end{tabular}
\label{table:relational}
\vspace{-20pt}
\end{table*}

\tpar{Pose and relation prediction}
To get the predicted pose and relations for future steps, we utilize a latent space dynamics model to propagate the current state information to the future after multiple actions.  
We define the latent space dynamics to propagate the $\glatent^{i}_{t}$ to ${\glatent^{\prime}}^{i}_{t+1}$ as ${\glatent^{\prime}}^{i}_{t+1} = \delta(\glatent^{i}_{t}, A_{t}(i, \theta))$.  
where $i$ is the ID of the object to manipulate and $\theta$ are continuous parameters. 
We use the same learned positional embedding to encode the discrete ID, as well as an MLP $M_\text{ac}$ to encode $\theta$. 
Since we have different skills, we use a different $\delta_{l}$ and $M_{\text{ac}_{l}}$ for each skill parameter $l$. 
After we get  ${\glatent^{\prime}}^{i}_{t+1}$, our framework decodes it to a predicted pose ${p^{\prime}}^{i}_{t+1}$ and a relation ${r^{\prime}}^{ij}_{t+1}$. 
Note we use  ${p^{\prime}}^{i}_{t+1}$ which discriminates from  $\hat{p}^{i}_{t+1}$, as ${p^{\prime}}^{i}_{t+1}$ comes from the observation $Z_t$ while $\hat{p}^{i}_{t+1}$ comes from the observation $Z_{t+1}$. 
With the ability to predict the pose and relation with latent space dynamics, even if there are occluded objects in $Z_{t+1}$, we can still estimate the pose of the occluded objects from $(Z_t, A_t, \delta)$. 
Finally, we can apply $\delta(\cdot)$ \(H\) times with a sequence of actions ${A_0, ..., A_{H-1}}$ to predict states $H$ time steps ahead as  ${p^{\prime}}^{i}_{t+H}$ and ${r^{\prime}}^{{ij}}_{t+H}$.

\tpar{Reasoning about occluded objects} 
We explore two different approaches to reason about occluded objects. Consider object \(k\) to be occluded. In our first approach, we use the predicted pose to transform the point cloud of the object \(k\) from time $t-1$, $O^{k}_{t-1}$, to $t$ in order to recover the missing observation as ${O^{\prime}}^{k}_{t}$. We then combine the transformed point cloud ${O^{\prime}}^{k}_{t}$ with the current observations as $(Z^{\prime}_t = {O^{1}_t, ..., {O^{\prime}}^{k}_{t}, ..., O^{N}_t})$. 
This allows the relational classifier to detect the relations at $t$ even if the \(k\)th object is completely occluded. We can repeat this process for any arbitrary number of occluded
objects. We name this \pca{} for \emph{Direct Occluded Object Memory}. %
Alternatively, we can copy the predicted latent space embedding ${\glatent^{\prime}}^{k}_{t}$ for the occluded object to the current latent state as $\glatent^{k}_{t}$ giving updated latent state $\glatentprimehat_{t} = \glatent^{0}_{t}, ... {\glatent^{\prime}}^{k}_{t}, ..., \glatent^{N}_{t}$. We call this approach \lsa{} for \emph{Latent Occluded Object Memory}.
Figure~\ref{fig:network_structure} illustrates \pca{} and \lsa{}. 

\tpar{Reasoning about reappeared objects}
When an occluded object reappears, the UVOS tracker identifies it as the previously occluded object. 
We can then remove the augmented memory state associated with the object from the model and instead pass this object's observation through the point cloud encoder like normal for both \pca{} and \lsa{}.

\tpar{Reasoning about novel objects' appearance}
UVOS identifies novel objects when they first appear. In this case, each  novel object segment 
receives a unique ID and is handled the same as any other observable object.

\tpar{Planning with \pca{} and \lsa{} }
For planning we use the cross-entropy method (CEM) as in~\cite{Huang-icra2023-graph-relations,huang2023latent}.
Since planning takes place prior to future observations, nothing changes when using \pca{} or \lsa{} compared to the model in~\cite{huang2023latent}, except that the initial state must encode any unobserved objects as described above.

\tpar{Training details}
Our training loss is the sum of three terms. 
First, we want our model to predict the pose, relations, and environment identity at the current step $t$. 
Given predictions $(\hat{p}^{i}_{t}, \hat{r}^{ij}_{t}, \hat{e}^{i}_t)$ and ground truth labels $(p^{i}_{t},r^{ij}_{t}, e^{i}_t)$, we get the current observation loss $L_c = \sum_{t=1}^{H}CE(\hat{r}^{ij}_{t}, r^{ij}_{t}) + ||\hat{p}^{i}_{t} - p^{i}_{t}||_2^{2} + CE(\hat{e}^{i}_t, e^{i}_t)$.  
Second, we require our models to learn that the predicted latent space from the previous observation ${\glatent^{\prime}}^{i}_{t+x}$ should be similar to the latent space from the current time observation $\glatent^{i}_{t+x}$. We call this the latent space regularization loss, defined as $L_{ls} = \sum_{t=1}^{H}\sum_{x=t}^{H}|| \glatent^{i}_{t+x}-{\glatent^{\prime}}^{i}_{t+x}||_2^{2}$. 
Last, our model should predict correct outputs from the state, ${\glatent^{\prime}}^{i}_{t+x}$ predicted by the latent dynamics; giving loss: $L_{d} = \sum_{t=1}^{H}\sum_{x=t}^{H}CE({{r^{\prime}}^{ij}_{t+x}}, r^{ij}_{t+x}) + ||{p^{\prime}}^{i}_{t+x} - p^{i}_{t+x}||_2^{2} + CE({e^{\prime}}^{i}_{t+x}, e^{i}_{t+x})$. 
We train on the full loss function $L = L_c + L_{ls} + L_d$ using the Adam optimizer.

\section{Experiments \& Results}\label{sec:experiments}
\begin{figure*}[t]
    \centering
    \includegraphics[width=0.98\columnwidth,clip,trim=0mm 0mm 0mm 0mm]{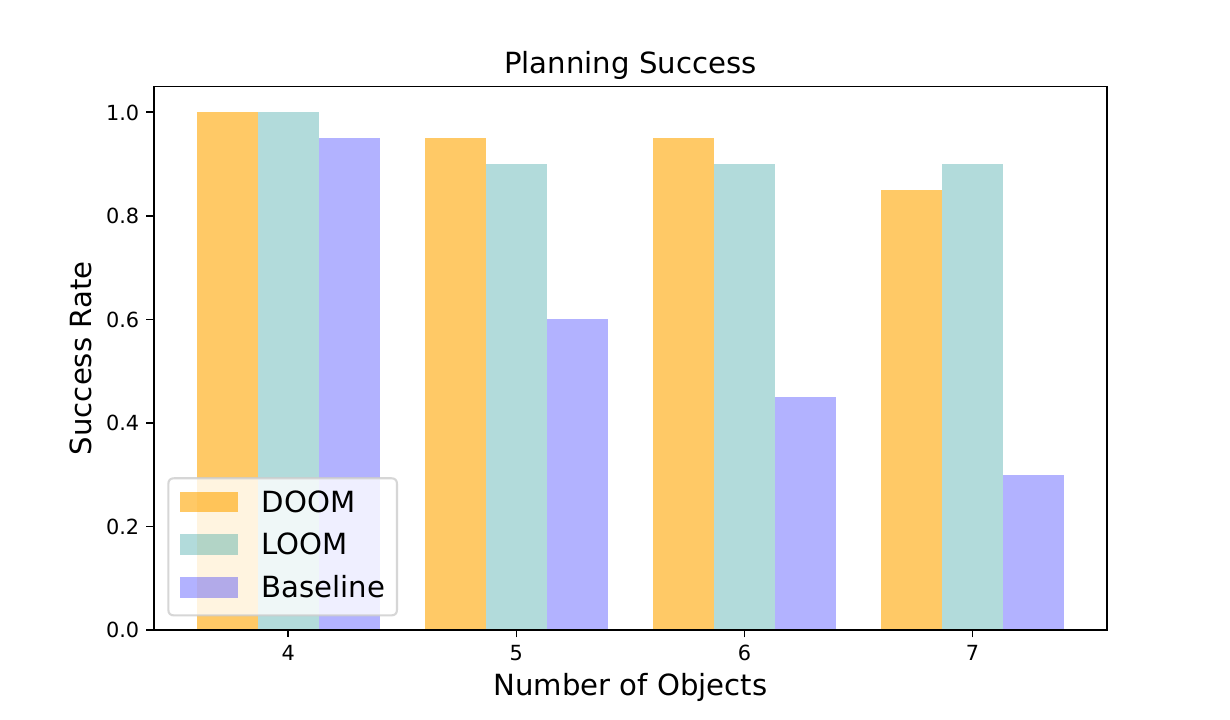} %
    \includegraphics[width=0.98\columnwidth,clip,trim=0mm 0mm 0mm 0mm]{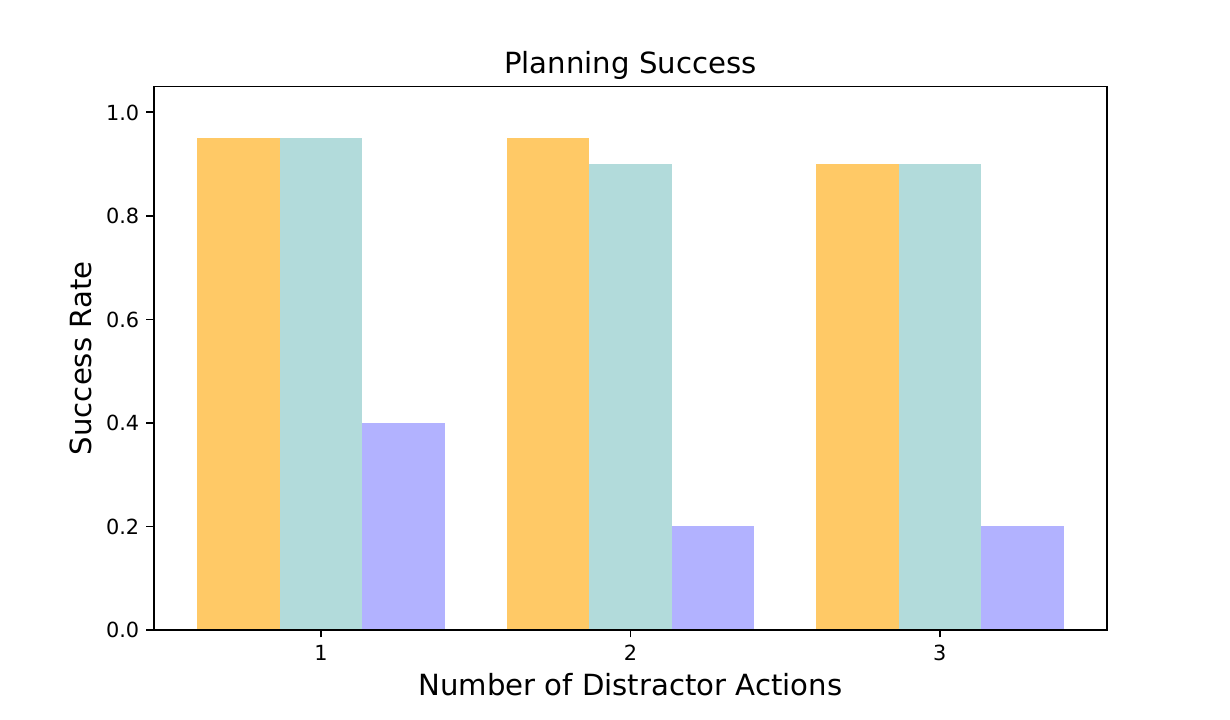}
    \vspace{-5pt}
    \caption{Execution success rate as a function of (left) the number of objects in the scene, and (right) the number of distractor actions.
    We find that our approaches perform consistently well across both conditions, outperforming the baseline by a large margin. This is especially prominent in terms of the number of distractor actions. The legend applies to both plots.}
    \label{fig:planning_success_comparison}
    \vspace{-10pt}
\end{figure*}

\begin{table*}[t]
\centering
\caption{Real World Planning Success.}
\begin{tabular}{ p{1.8cm}p{0.8cm}p{0.8cm}p{0.8cm}p{0.8cm} p{0.8cm} p{1.8cm}p{0.8cm}p{0.8cm}p{0.8cm} p{0.8cm}}
 \hline
 Objects & 4 & 5 & 6 & 7 & all & Distractors & 1 & 2 & 3 & all\\
 \hline
 \pca{}   & 5/5 & 5/5 & 5/5 & 4/5 & \textbf{19/20} & \pca{} & 5/5 & 4/5 & 5/5 & \textbf{14/15} \\
 \lsa{}     & 5/5 & 4/5 & 5/5 & 5/5 & \textbf{19/20} & \lsa{}& 4/5 & 5/5 & 4/5 & \textbf{13/15} \\
  \cite{Huang-icra2023-graph-relations, huang2023latent}     & - & - & - & - & 0/20 & \cite{Huang-icra2023-graph-relations, huang2023latent} & - & - & - & 0/15 \\
 \hline
\end{tabular}
\label{table:real}
\vspace{-20pt}
\end{table*}

\subsection{Dataset Generation and Relations Definition}
We generate a dataset with more than 20,000 skill executions.
We generate skill executions with three steps:
(1): We create scenes with a variable number of objects in the Isaac Gym simulator~\cite{isaacgym}.
(2): We generate a random skill primitive with random continuous parameters.
(3): We execute this skill in simulation.
We repeat this process to create a training sequence with multiple timesteps.
After the skill executions, we record the point clouds, poses, bounding boxes, and relations pre and post-manipulation. During data collection, if an object $O^k_{t}$ disappears at a timestep $t$ due to occlusion, we augment the point clouds at $t$ with the transformed point cloud of the disappeared object from a previous timestep. We show an example of the dataset generation process in Fig.~\ref{fig:simulation_figure}.
In this section, the number of objects is defined as the number of segments including object and environment segments. 

We define 9 relations for model evaluation: \textit{left, right, front, behind, above, below, contact, boundary, inside}.
The relations are defined based on the pose and extent of object bounding boxes in the simulation.
The simulator provides us with the \textit{contact} relation.
We define inside(A, B) = 1 if the bounding box of A is surrounded by the bounding box of B, except that the inside object can stick out the top of the larger object a maximum of 3cm.
We define the boundary relation as in~\cite{huang2023latent} and other relations (left, right, etc.) as in~\cite{paxton-corl2021-semantic-placement}.
Furthermore, we use the \textit{inside} relation as a heuristic during planning and don't attempt to directly grasp or push objects predicted to be contained inside others. %

\subsection{Baseline Approach}
Our framework is built upon previous works~\cite{Huang-icra2023-graph-relations, huang2023latent}, which  include comparisons to different baselines and intensive ablation studies. We refer the readers to our prior works to see the effectiveness of our planning framework compared to other alternatives, such as the one mainly relying on explicit pose estimation similar to \cite{xu2020learning}. %
In this paper, we use a transformer to implicitly encode the history similar to~\cite{bonatti2022pact} as our baseline approach. The transformer learns how to combine the history of previous observations and the current observation. 
For each observation $Z_t$, we concatenate the accumulated history with a type embedding $h_{type}$ and feed it into an MLP to get $h_t = MLP([E_p(E(X_t)) + h_{t-1}, h_{type}(t)])$ while we use $h_0 = E_p(E(X_0))$. The type embedding $h_{type}$ encodes whether the segment is 
observable at the current step. 
We pass $h_t$ to the transformer to model the interaction between object tokens, and then use the decoders to predict object relations and poses.

\vspace{-2pt}
\subsection{Relational Prediction Evaluation}
To test our hypothesis that an explicit object-oriented memory representation is more effective, we evaluate the ability of \pca{} and \lsa{} to predict relations after an action.
We define a distractor action as an action applied after some objects disappear from the scene due to occlusion. After an object disappears from the scene, we manipulate other objects several times and then ask the model to reason about the occluded object.
In the evaluation across different numbers of objects shown in Table.~\ref{table:relational}, \pca{}/\lsa{} improve over \transa{} by close to $20\%$.  %
In the evaluation across different numbers of distractor actions shown in Table.~\ref{table:relational}, \pca{}/\lsa{} improve over \transa{} by more than $10\%$. %

\vspace{-2pt}
\subsection{Planning to Goal Relations Success Rate Evaluation}
\subsubsection{Generalization to a different number of objects}
We first show how well our approaches generalize to a variable number of objects. Examples in our training dataset contain a maximum of 5 objects. 
In the qualitative results shown in Fig.~\ref{fig:simulation_figure}, we show that our approaches can generalize to 8 novel objects with different shapes and different viewpoints.
In the results shown in Fig.~\ref{fig:planning_success_comparison}, we find that our approaches \lsa{} and \pca{} perform well and are robust to changes in the number of objects presented in the scene.
In contrast, the baseline approach performs well on scenes with 4 objects, but struggles on scenes with more objects.

\subsubsection{Generalization to a different number of distractor actions}
\begin{figure}[t]
    \centering
    \includegraphics[width=1\columnwidth]{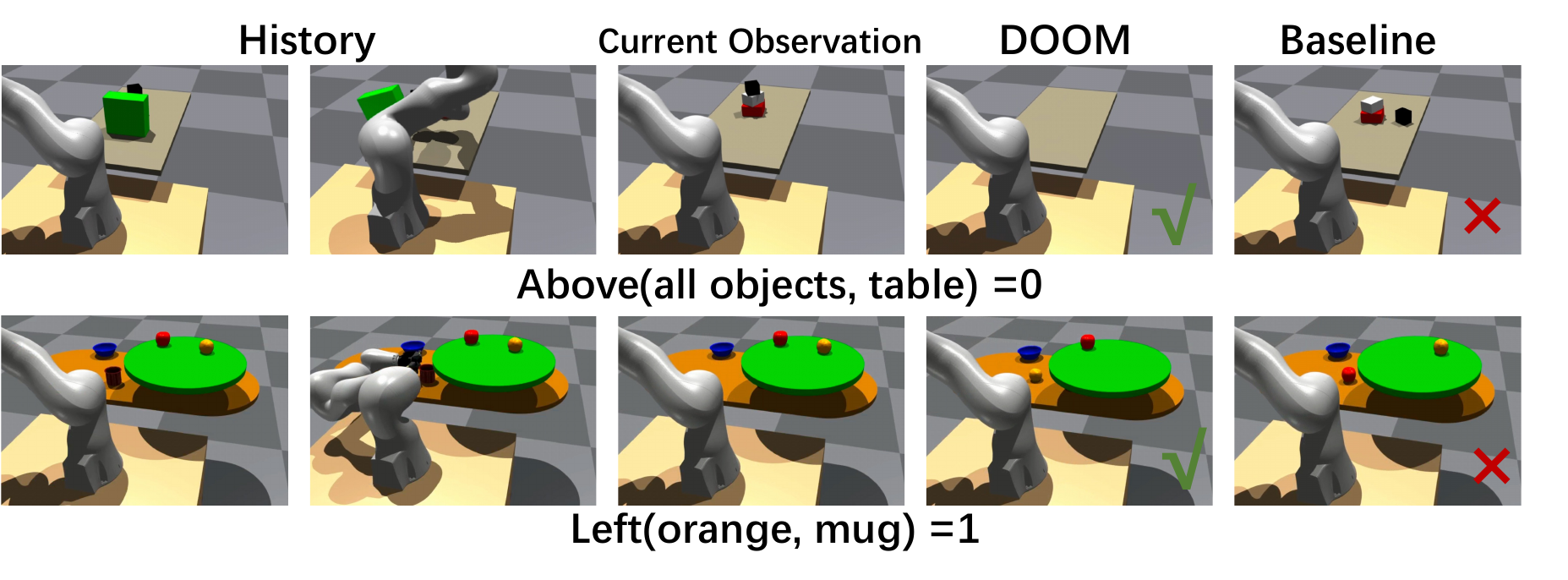}
    \vspace{-15pt}
    \caption{We show two failure cases of the baseline, where our approach achieves the goal relations. For the first example, after the robot pushes the red box off the table, two novel objects appear in the current observation. The goal is to remove all the objects on the table. \pca{} achieves the goal by pushing the red object while \transa{} fails because it pushes a wrong (black) object. For the second example, after the robot pushes the mug, the mug is occluded by the shelf. To achieve the goal, \pca{} picks and places the orange while \transa{} picks and places a wrong object~(apple).}
    \label{fig:failure}\vspace{-21pt}
\end{figure}

The advantage of explicit object-oriented memory is prominent when generalizing to different number of distractor actions~(Fig.~\ref{fig:planning_success_comparison}). Our approaches perform well for different numbers of distractor actions, but the baseline performance drops significantly with even one distractor action.
We show qualitative results in Fig.~\ref{fig:failure}.
Note that distractor actions are not generated during training. %
This verifies our hypothesis that the performance of the implicit memory approach can significantly decrease with distractor actions.
Note that the difference in the planning success rate is more significant than the F1 score of the relational prediction. 
That is because when evaluating the relational predictions, all relations across all objects were considered, including visible objects, so that \transa{} performs reasonably well.
However, when evaluating planning success, the goal relations include occluded objects so \transa{} performs poorly, as it does not reason about occluded objects well. 

\subsubsection{Real-world planning success evaluation}
Since the \transa{} performs poorly in simulation, we only show our approaches \pca{} and \lsa{} in the real-world evaluation.
We ran 35 trials per approach with different numbers of objects and numbers of distractor actions. 
In the results shown in Table~\ref{table:real}, we find our approaches perform well for different numbers of objects and distractor actions. 
Note that prior works \cite{Huang-icra2023-graph-relations, huang2023latent} would achieve a 0\% success rate on this evaluation because of occluded objects.

\vspace{-2pt}
\subsection{Qualitative Analysis}\label{sec:quantitative}%
We visualize executions of our approaches in Fig.~1. 
The first example shows how our framework can understand occluded objects and multiple objects with the same appearance.
The robot first pushes the yellow mustard to below the shelf.
Then the robot pushes the grey cup that is above the shelf as a distractor action.
After these two actions, our model can predict the pose of the occluded yellow mustard.
The robot can pick and place the blue cup above the shelf to achieve the goal relation.
For the second example, after the robot puts the apple inside the pitcher, the apple is occluded.
Then we make the robot move the pitcher as a distractor action.
After the distractor action  our framework can still remember the location of the apple and can pick and place the orange inside the pitcher to make it in contact with the apple to meet the goal.
For the third example, after the robot picks and places the green box inside the pitcher and moves the pitcher, our framework can understand where the green box is.
Then the robot chooses to pick the pitcher and dump it to make the green box in contact with the table.
For the fourth example, after the robot pushes the Cheez-It box, two novel objects appear. Our model understands that these are new objects and the Cheez-It box has disappeared, so the robot can achieve the goal of pushing all the objects to the boundary of the table.

\section{Conclusion}\label{sec:discussion}%
We examined how to encode memory to reason about occluded objects, the appearance of novel objects, and object reappearance. 
We propose two approaches \pca{} and \lsa{} to incorporate memory into transformer-based relational dynamics. Through comparison to a baseline implicit memory model, we verify that an explicit object-oriented memory is better at generalizing in terms of number of objects and number of distractor actions. 
Our work is the first to jointly leverage a multi-object, relational dynamics model and a UVOS object discovery and tracking model to reason about occluded objects. It shows the promise of utilizing object discovery models for long-term manipulation tasks.

We identify several areas for future research.
One promising idea is to integrate the tracker and planner and train a full system end-to-end.
For example, we could leverage the pose prediction from \pca{} as a prior for the tracker.
We would additionally like to incorporate a mobile base requiring more complicated memory management and examine longer-horizon planning tasks.

\section*{Acknowledgments}
The authors thank Parker Ewen and Griffin Tabor for the useful discussion. 
This work was partially supported by NSF Awards \#2024778 and \#1911232, by DARPA under grant \mbox{N66001-19-2-4035}, by a Sloan Research Fellowship, by ODA award \mbox{ODA23008GR}, and by ONR awards \mbox{N00024-10-D-6318} and \mbox{N0014-21-1-2052}.

\bibliographystyle{IEEEtran}
\bibliography{references}

\begin{thebibliography}{10}
\providecommand{\url}[1]{#1}
\csname url@samestyle\endcsname
\providecommand{\newblock}{\relax}
\providecommand{\bibinfo}[2]{#2}
\providecommand{\BIBentrySTDinterwordspacing}{\spaceskip=0pt\relax}
\providecommand{\BIBentryALTinterwordstretchfactor}{4}
\providecommand{\BIBentryALTinterwordspacing}{\spaceskip=\fontdimen2\font plus
\BIBentryALTinterwordstretchfactor\fontdimen3\font minus \fontdimen4\font\relax}
\providecommand{\BIBforeignlanguage}[2]{{%
\expandafter\ifx\csname l@#1\endcsname\relax
\typeout{** WARNING: IEEEtran.bst: No hyphenation pattern has been}%
\typeout{** loaded for the language `#1'. Using the pattern for}%
\typeout{** the default language instead.}%
\else
\language=\csname l@#1\endcsname
\fi
#2}}
\providecommand{\BIBdecl}{\relax}
\BIBdecl

\bibitem{Moravec_1988}
\BIBentryALTinterwordspacing
H.~P. Moravec, ``Sensor fusion in certainty grids for mobile robots,'' \emph{AI Magazine}, vol.~9, no.~2, p.~61, Jun. 1988. [Online]. Available: \url{https://ojs.aaai.org/aimagazine/index.php/aimagazine/article/view/676}
\BIBentrySTDinterwordspacing

\bibitem{leonard-ijrr1992}
\BIBentryALTinterwordspacing
J.~J. Leonard, H.~F. Durrant-Whyte, and I.~J. Cox, ``Dynamic map building for an autonomous mobile robot,'' \emph{The International Journal of Robotics Research}, vol.~11, no.~4, pp. 286--298, 1992. [Online]. Available: \url{https://doi.org/10.1177/027836499201100402}
\BIBentrySTDinterwordspacing

\bibitem{Kuipers-2000-ssh}
\BIBentryALTinterwordspacing
B.~Kuipers, ``The spatial semantic hierarchy,'' \emph{Artificial Intelligence}, vol. 119, no.~1, pp. 191--233, 2000. [Online]. Available: \url{https://www.sciencedirect.com/science/article/pii/S0004370200000175}
\BIBentrySTDinterwordspacing

\bibitem{chown-1999-r-plan}
\BIBentryALTinterwordspacing
E.~Chown, ``Making predictions in an uncertain world: Environmental structure and cognitive maps,'' \emph{Adaptive Behavior}, vol.~7, no.~1, pp. 17--33, 1999. [Online]. Available: \url{https://doi.org/10.1177/105971239900700102}
\BIBentrySTDinterwordspacing

\bibitem{cadena-tro2016-slam}
C.~Cadena, L.~Carlone, H.~Carrillo, Y.~Latif, D.~Scaramuzza, J.~Neira, I.~Reid, and J.~J. Leonard, ``Past, present, and future of simultaneous localization and mapping: Toward the robust-perception age,'' \emph{IEEE Transactions on Robotics}, vol.~32, no.~6, pp. 1309--1332, 2016.

\bibitem{Pillai-RSS-15}
S.~Pillai and J.~Leonard, ``Monocular slam supported object recognition,'' in \emph{Proceedings of Robotics: Science and Systems}, Rome, Italy, July 2015.

\bibitem{herbst-icra2014}
E.~Herbst, P.~Henry, and D.~Fox, ``Toward online 3-d object segmentation and mapping,'' in \emph{IEEE International Conference on Robotics and Automation (ICRA)}, 2014, pp. 3193--3200.

\bibitem{faulhammer-ral2017}
T.~Fäulhammer, R.~Ambruş, C.~Burbridge, M.~Zillich, J.~Folkesson, N.~Hawes, P.~Jensfelt, and M.~Vincze, ``Autonomous learning of object models on a mobile robot,'' \emph{IEEE Robotics and Automation Letters}, vol.~2, no.~1, pp. 26--33, 2017.

\bibitem{weiyu-rss2023}
\BIBentryALTinterwordspacing
W.~Liu, Y.~Du, T.~Hermans, S.~Chernova, and C.~Paxton, ``{StructDiffusion: Language-Guided Creation of Physically-Valid Structures using Unseen Objects},'' in \emph{Robotics: Science and Systems}, 2023. [Online]. Available: \url{https://structdiffusion.github.io/}
\BIBentrySTDinterwordspacing

\bibitem{zhu2020hierarchical}
\BIBentryALTinterwordspacing
Y.~Zhu, J.~Tremblay, S.~Birchfield, and Y.~Zhu, ``Hierarchical planning for long-horizon manipulation with geometric and symbolic scene graphs,'' in \emph{IEEE International Conference on Robotics and Automation (ICRA)}, 2021. [Online]. Available: \url{https://arxiv.org/abs/2012.07277}
\BIBentrySTDinterwordspacing

\bibitem{lin2022efficient}
\BIBentryALTinterwordspacing
Y.~Lin, A.~S. Wang, E.~Undersander, and A.~Rai, ``Efficient and interpretable robot manipulation with graph neural networks,'' \emph{{IEEE Robotics and Automation Letters}}, 2022. [Online]. Available: \url{https://arxiv.org/abs/2102.13177}
\BIBentrySTDinterwordspacing

\bibitem{sharma2020relational}
\BIBentryALTinterwordspacing
M.~Sharma and O.~Kroemer, ``Relational learning for skill preconditions,'' in \emph{Conference on Robot Learning (CoRL)}, 2020. [Online]. Available: \url{https://arxiv.org/abs/2012.01693}
\BIBentrySTDinterwordspacing

\bibitem{liu-icra2022-structformer}
\BIBentryALTinterwordspacing
W.~Liu, C.~Paxton, T.~Hermans, and D.~Fox, ``{StructFormer: Learning Spatial Structure for Language-Guided Semantic Rearrangement of Novel Objects},'' in \emph{IEEE International Conference on Robotics and Automation (ICRA)}, 2022. [Online]. Available: \url{https://sites.google.com/view/structformer}
\BIBentrySTDinterwordspacing

\bibitem{murali20206}
\BIBentryALTinterwordspacing
A.~Murali, A.~Mousavian, C.~Eppner, C.~Paxton, and D.~Fox, ``6-dof grasping for target-driven object manipulation in clutter,'' in \emph{IEEE International Conference on Robotics and Automation (ICRA)}.\hskip 1em plus 0.5em minus 0.4em\relax IEEE, 2020, pp. 6232--6238. [Online]. Available: \url{https://arxiv.org/abs/1912.03628}
\BIBentrySTDinterwordspacing

\bibitem{paxton-corl2021-semantic-placement}
\BIBentryALTinterwordspacing
C.~Paxton, C.~Xie, T.~Hermans, and D.~Fox, ``{Predicting Stable Configurations for Semantic Placement of Novel Objects},'' in \emph{Conference on Robot Learning (CoRL)}, 11 2021. [Online]. Available: \url{https://arxiv.org/abs/2108.12062}
\BIBentrySTDinterwordspacing

\bibitem{Qureshi-RSS-21}
\BIBentryALTinterwordspacing
A.~H. Qureshi, A.~Mousavian, C.~Paxton, M.~Yip, and D.~Fox, ``{NeRP: Neural Rearrangement Planning for Unknown Objects},'' in \emph{Proceedings of Robotics: Science and Systems}, Virtual, July 2021. [Online]. Available: \url{https://arxiv.org/abs/2106.01352}
\BIBentrySTDinterwordspacing

\bibitem{Huang-icra2023-graph-relations}
\BIBentryALTinterwordspacing
Y.~Huang, A.~Conkey, and T.~Hermans, ``{Planning for Multi-Object Manipulation with Graph Neural Network Relational Classifiers},'' in \emph{IEEE International Conference on Robotics and Automation (ICRA)}, 2023. [Online]. Available: \url{https://arxiv.org/abs/2209.11943}
\BIBentrySTDinterwordspacing

\bibitem{huang2023latent}
Y.~Huang, N.~C. Taylor, A.~Conkey, W.~Liu, and T.~Hermans, ``Latent space planning for multi-object manipulation with environment-aware relational classifiers,'' \emph{arXiv preprint arXiv:2305.10857}, 2023.

\bibitem{bonatti2022pact}
R.~Bonatti, S.~Vemprala, S.~Ma, F.~Frujeri, S.~Chen, and A.~Kapoor, ``Pact: Perception-action causal transformer for autoregressive robotics pre-training,'' \emph{arXiv preprint arXiv:2209.11133}, 2022.

\bibitem{wani2020multion}
S.~Wani, S.~Patel, U.~Jain, A.~X. Chang, and M.~Savva, ``Multi-on: Benchmarking semantic map memory using multi-object navigation,'' in \emph{Neural Information Processing Systems (NeurIPS)}, 2020.

\bibitem{cartillier2021semantic}
V.~Cartillier, Z.~Ren, N.~Jain, S.~Lee, I.~Essa, and D.~Batra, ``Semantic mapnet: Building allocentric semantic maps and representations from egocentric views,'' in \emph{Proceedings of the AAAI Conference on Artificial Intelligence}, vol.~35, no.~2, 2021, pp. 964--972.

\bibitem{ebert2017self}
F.~Ebert, C.~Finn, A.~X. Lee, and S.~Levine, ``Self-supervised visual planning with temporal skip connections.'' \emph{CoRL}, vol.~12, p.~16, 2017.

\bibitem{kim2022memory}
H.~Kim, Y.~Ohmura, and Y.~Kuniyoshi, ``Memory-based gaze prediction in deep imitation learning for robot manipulation,'' in \emph{2022 International Conference on Robotics and Automation (ICRA)}.\hskip 1em plus 0.5em minus 0.4em\relax IEEE, 2022, pp. 2427--2433.

\bibitem{Caelles_arXiv_2019}
S.~Caelles, J.~Pont-Tuset, , F.~Perazzi, A.~Montes, K.-K. Maninis, and L.~{Van Gool}, ``The 2019 davis challenge on vos: Unsupervised multi-object segmentation,'' \emph{arXiv:1905.00737}, 2019.

\bibitem{luiten2020unovost}
J.~Luiten, I.~E. Zulfikar, and B.~Leibe, ``Unovost: Unsupervised offline video object segmentation and tracking,'' in \emph{Proceedings of the IEEE/CVF Winter Conference on Applications of Computer Vision}, 2020, pp. 2000--2009.

\bibitem{Lin_2021_ICCV}
H.~Lin, R.~Wu, S.~Liu, J.~Lu, and J.~Jia, ``Video instance segmentation with a propose-reduce paradigm,'' in \emph{Proceedings of the IEEE/CVF International Conference on Computer Vision (ICCV)}, October 2021, pp. 1739--1748.

\bibitem{Zhou_2021_CVPR}
T.~Zhou, J.~Li, X.~Li, and L.~Shao, ``Target-aware object discovery and association for unsupervised video multi-object segmentation,'' in \emph{Proceedings of the IEEE/CVF Conference on Computer Vision and Pattern Recognition (CVPR)}, June 2021, pp. 6985--6994.

\bibitem{yuan2023maximal}
J.~Yuan, J.~Patravali, H.~Nguyen, C.~Kim, and L.~Fuxin, ``Maximal cliques on multi-frame proposal graph for unsupervised video object segmentation,'' \emph{arXiv preprint arXiv:2301.12352}, 2023.

\bibitem{garrett2020pddlstream}
\BIBentryALTinterwordspacing
C.~R. Garrett, T.~Lozano-P{\'e}rez, and L.~P. Kaelbling, ``Pddlstream: Integrating symbolic planners and blackbox samplers via optimistic adaptive planning,'' in \emph{Proceedings of the International Conference on Automated Planning and Scheduling}, vol.~30, 2020, pp. 440--448. [Online]. Available: \url{https://arxiv.org/abs/1802.08705}
\BIBentrySTDinterwordspacing

\bibitem{li2020towards}
\BIBentryALTinterwordspacing
R.~Li, A.~Jabri, T.~Darrell, and P.~Agrawal, ``Towards practical multi-object manipulation using relational reinforcement learning,'' in \emph{IEEE International Conference on Robotics and Automation (ICRA)}, 2020, pp. 4051--4058. [Online]. Available: \url{https://arxiv.org/abs/1912.11032}
\BIBentrySTDinterwordspacing

\bibitem{xu2020learning}
Z.~Xu, Z.~He, J.~Wu, and S.~Song, ``Learning 3d dynamic scene representations for robot manipulation,'' in \emph{Conference on Robot Learning (CoRL)}, 2020.

\bibitem{du2022play}
M.~Du, O.~Y. Lee, S.~Nair, and C.~Finn, ``Play it by ear: Learning skills amidst occlusion through audio-visual imitation learning,'' \emph{arXiv preprint arXiv:2205.14850}, 2022.

\bibitem{curtis2022long}
A.~Curtis, X.~Fang, L.~P. Kaelbling, T.~Lozano-P{\'e}rez, and C.~R. Garrett, ``Long-horizon manipulation of unknown objects via task and motion planning with estimated affordances,'' in \emph{2022 International Conference on Robotics and Automation (ICRA)}.\hskip 1em plus 0.5em minus 0.4em\relax IEEE, 2022, pp. 1940--1946.

\bibitem{shafiullah2022clip}
N.~M.~M. Shafiullah, C.~Paxton, L.~Pinto, S.~Chintala, and A.~Szlam, ``Clip-fields: Weakly supervised semantic fields for robotic memory,'' \emph{arXiv preprint arXiv:2210.05663}, 2022.

\bibitem{jockel2009sparse}
S.~Jockel, F.~Lindner, and J.~Zhang, ``Sparse distributed memory for experience-based robot manipulation,'' in \emph{2008 IEEE International Conference on Robotics and Biomimetics}.\hskip 1em plus 0.5em minus 0.4em\relax IEEE, 2009, pp. 1298--1303.

\bibitem{jockel2009robot}
S.~Jockel, M.~Mendes, J.~Zhang, A.~P. Coimbra, and M.~Cris{\'o}stomo, ``Robot navigation and manipulation based on a predictive associative memory,'' in \emph{2009 IEEE 8th International Conference on Development and Learning}.\hskip 1em plus 0.5em minus 0.4em\relax IEEE, 2009, pp. 1--7.

\bibitem{bessler2018knowledge}
D.~Be{\ss}ler, S.~Koralewski, and M.~Beetz, ``Knowledge representation for cognition-and learning-enabled robot manipulation.'' in \emph{CogRob@ KR}, 2018, pp. 11--19.

\bibitem{Cai_2022_CVPR}
J.~Cai, M.~Xu, W.~Li, Y.~Xiong, W.~Xia, Z.~Tu, and S.~Soatto, ``Memot: Multi-object tracking with memory,'' in \emph{Proceedings of the IEEE/CVF Conference on Computer Vision and Pattern Recognition (CVPR)}, June 2022, pp. 8090--8100.

\bibitem{Fang2017RecurrentAN}
K.~Fang, Y.~Xiang, X.~Li, and S.~Savarese, ``Recurrent autoregressive networks for online multi-object tracking,'' \emph{2018 IEEE Winter Conference on Applications of Computer Vision (WACV)}, 2017.

\bibitem{cheng2022xmem}
H.~K. Cheng and A.~G. Schwing, ``{XMem}: Long-term video object segmentation with an atkinson-shiffrin memory model,'' in \emph{ECCV}, 2022.

\bibitem{Oh_2019_ICCV}
S.~W. Oh, J.-Y. Lee, N.~Xu, and S.~J. Kim, ``Video object segmentation using space-time memory networks,'' in \emph{Proceedings of the IEEE/CVF International Conference on Computer Vision (ICCV)}, October 2019.

\bibitem{di2023one}
F.~Di~Felice, S.~D'Avella, A.~Remus, P.~Tripicchio, and C.~A. Avizzano, ``One-shot imitation learning with graph neural networks for pick-and-place manipulation tasks,'' \emph{IEEE Robotics and Automation Letters}, 2023.

\bibitem{chen2023predicting}
\BIBentryALTinterwordspacing
H.~Chen, Y.~Niu, K.~Hong, S.~Liu, Y.~Wang, Y.~Li, and K.~R. Driggs-Campbell, ``Predicting object interactions with behavior primitives: An application in stowing tasks,'' in \emph{7th Annual Conference on Robot Learning}, 2023. [Online]. Available: \url{https://openreview.net/forum?id=VH6WIPF4Sj}
\BIBentrySTDinterwordspacing

\bibitem{kulshrestha2023structural}
\BIBentryALTinterwordspacing
M.~Kulshrestha and A.~H. Qureshi, ``Structural concept learning via graph attention for multi-level rearrangement planning,'' in \emph{7th Annual Conference on Robot Learning}, 2023. [Online]. Available: \url{https://openreview.net/forum?id=D0X97ODIYK}
\BIBentrySTDinterwordspacing

\bibitem{yuan2022sornet}
\BIBentryALTinterwordspacing
W.~Yuan, C.~Paxton, K.~Desingh, and D.~Fox, ``Sornet: Spatial object-centric representations for sequential manipulation,'' in \emph{Conference on Robot Learning (CoRL)}.\hskip 1em plus 0.5em minus 0.4em\relax PMLR, 2022, pp. 148--157. [Online]. Available: \url{https://openreview.net/forum?id=mOLu2rODIJF}
\BIBentrySTDinterwordspacing

\bibitem{shridhar2023perceiver}
M.~Shridhar, L.~Manuelli, and D.~Fox, ``Perceiver-actor: A multi-task transformer for robotic manipulation,'' in \emph{Conference on Robot Learning}.\hskip 1em plus 0.5em minus 0.4em\relax PMLR, 2023, pp. 785--799.

\bibitem{zitkovich2023rt}
\BIBentryALTinterwordspacing
B.~Zitkovich, T.~Yu, S.~Xu, P.~Xu, T.~Xiao, F.~Xia, J.~Wu, P.~Wohlhart, S.~Welker, A.~Wahid, quan vuong, V.~Vanhoucke, H.~Tran, R.~Soricut, A.~Singh, J.~Singh, P.~Sermanet, P.~R. Sanketi, G.~Salazar, M.~S. Ryoo, K.~Reymann, K.~Rao, K.~Pertsch, I.~Mordatch, H.~Michalewski, Y.~Lu, S.~Levine, L.~Lee, T.-W.~E. Lee, I.~Leal, Y.~Kuang, D.~Kalashnikov, R.~Julian, N.~J. Joshi, A.~Irpan, brian ichter, J.~Hsu, A.~Herzog, K.~Hausman, K.~Gopalakrishnan, C.~Fu, P.~Florence, C.~Finn, K.~A. Dubey, D.~Driess, T.~Ding, K.~M. Choromanski, X.~Chen, Y.~Chebotar, J.~Carbajal, N.~Brown, A.~Brohan, M.~G. Arenas, and K.~Han, ``{RT}-2: Vision-language-action models transfer web knowledge to robotic control,'' in \emph{7th Annual Conference on Robot Learning}, 2023. [Online]. Available: \url{https://openreview.net/forum?id=XMQgwiJ7KSX}
\BIBentrySTDinterwordspacing

\bibitem{simeonov2023shelving}
\BIBentryALTinterwordspacing
A.~Simeonov, A.~Goyal, L.~Manuelli, Y.-C. Lin, A.~Sarmiento, A.~R. Garcia, P.~Agrawal, and D.~Fox, ``Shelving, stacking, hanging: Relational pose diffusion for multi-modal rearrangement,'' in \emph{7th Annual Conference on Robot Learning}, 2023. [Online]. Available: \url{https://openreview.net/forum?id=_xFJuqBId8c}
\BIBentrySTDinterwordspacing

\bibitem{chi2023diffusionpolicy}
C.~Chi, S.~Feng, Y.~Du, Z.~Xu, E.~Cousineau, B.~Burchfiel, and S.~Song, ``Diffusion policy: Visuomotor policy learning via action diffusion,'' in \emph{Proceedings of Robotics: Science and Systems (RSS)}, 2023.

\bibitem{kim2020learning}
\BIBentryALTinterwordspacing
B.~Kim and L.~Shimanuki, ``Learning value functions with relational state representations for guiding task-and-motion planning,'' in \emph{Conference on Robot Learning (CoRL)}, 2019. [Online]. Available: \url{http://people.csail.mit.edu/beomjoon/publications/kim-corl19.pdf}
\BIBentrySTDinterwordspacing

\bibitem{driess2020deep}
\BIBentryALTinterwordspacing
D.~Driess, J.-S. Ha, and M.~Toussaint, ``Deep visual reasoning: Learning to predict action sequences for task and motion planning from an initial scene image,'' in \emph{Proceedings of Robotics: Science and Systems}, 2020. [Online]. Available: \url{https://arxiv.org/abs/2006.05398}
\BIBentrySTDinterwordspacing

\bibitem{garrett-icra2020}
C.~R. Garrett, C.~Paxton, T.~Lozano-Pérez, L.~P. Kaelbling, and D.~Fox, ``Online replanning in belief space for partially observable task and motion problems,'' in \emph{IEEE International Conference on Robotics and Automation (ICRA)}, 2020, pp. 5678--5684.

\bibitem{liang-icra2022}
\BIBentryALTinterwordspacing
J.~Liang, M.~Sharma, A.~LaGrassa, S.~Vats, S.~Saxena, and O.~Kroemer, ``{Search-Based Task Planning with Learned Skill Effect Models for Lifelong Robotic Manipulation},'' in \emph{IEEE International Conference on Robotics and Automation (ICRA)}, 2022. [Online]. Available: \url{https://arxiv.org/abs/2109.08771}
\BIBentrySTDinterwordspacing

\bibitem{yang2022sequence}
Z.~Yang, C.~R. Garrett, and D.~Fox, ``Sequence-based plan feasibility prediction for efficient task and motion planning,'' \emph{arXiv preprint arXiv:2211.01576}, 2022.

\bibitem{xiang2017posecnn}
Y.~Xiang, T.~Schmidt, V.~Narayanan, and D.~Fox, ``Posecnn: A convolutional neural network for 6d object pose estimation in cluttered scenes,'' \emph{arXiv preprint arXiv:1711.00199}, 2017.

\bibitem{fischler1981random}
M.~A. Fischler and R.~C. Bolles, ``Random sample consensus: a paradigm for model fitting with applications to image analysis and automated cartography,'' \emph{Communications of the ACM}, vol.~24, no.~6, pp. 381--395, 1981.

\bibitem{wu2019pointconv}
\BIBentryALTinterwordspacing
W.~Wu, Z.~Qi, and L.~Fuxin, ``{PointConv: Deep Convolutional Networks on 3D Point Clouds},'' in \emph{Proceedings of the IEEE/CVF Conference on Computer Vision and Pattern Recognition (CVPR)}, 2019, pp. 9621--9630. [Online]. Available: \url{https://arxiv.org/abs/1811.07246}
\BIBentrySTDinterwordspacing

\bibitem{cui2022positional}
\BIBentryALTinterwordspacing
H.~Cui, Z.~Lu, P.~Li, and C.~Yang, ``On positional and structural node features for graph neural networks on non-attributed graphs,'' in \emph{Proceedings of the 31st ACM International Conference on Information \& Knowledge Management}, 2022, pp. 3898--3902. [Online]. Available: \url{https://arxiv.org/abs/2107.01495}
\BIBentrySTDinterwordspacing

\bibitem{isaacgym}
\BIBentryALTinterwordspacing
V.~Makoviychuk, L.~Wawrzyniak, Y.~Guo, M.~Lu, K.~Storey, M.~Macklin, D.~Hoeller, N.~Rudin, A.~Allshire, A.~Handa \emph{et~al.}, ``Isaac gym: High performance gpu-based physics simulation for robot learning,'' in \emph{Advances in Neural Information Processing Systems}, 2021. [Online]. Available: \url{https://sites.google.com/view/isaacgym-nvidia}
\BIBentrySTDinterwordspacing

\bibitem{vaswani2017attention}
A.~Vaswani, N.~Shazeer, N.~Parmar, J.~Uszkoreit, L.~Jones, A.~N. Gomez, {\L}.~Kaiser, and I.~Polosukhin, ``Attention is all you need,'' \emph{Advances in neural information processing systems}, vol.~30, 2017.

\bibitem{wang2020solov2}
X.~Wang, R.~Zhang, T.~Kong, L.~Li, and C.~Shen, ``Solov2: Dynamic and fast instance segmentation,'' \emph{Advances in Neural information processing systems}, vol.~33, pp. 17\,721--17\,732, 2020.

\bibitem{oh2019video}
S.~W. Oh, J.-Y. Lee, N.~Xu, and S.~J. Kim, ``Video object segmentation using space-time memory networks,'' in \emph{Proceedings of the IEEE/CVF International Conference on Computer Vision}, 2019, pp. 9226--9235.

\end{thebibliography}

\newpage
\pagenumbering{arabic}%
\renewcommand*{\thepage}{A\arabic{page}} %
\appendix
\section{Appendix}
\subsection{Preliminary Knowledge}

We provide the background information here about the transformers, which is the core component of our approaches. 
Transformers, an attention-based model, is proposed in ~\cite{vaswani2017attention} for sequential data.
The input for a transformer is a sequence $S$ with length k and the output is another sequence $S^{\prime}$ with the same length k. 
The heart of the transformer is the attention between different nodes in the sequence $S$.  
The specific attention used in a transformer is called "Scaled Dot-Product Attention". 
Each input feature is projected to a query, key, and value. 
The specific function to compute attention is shown in equation.~\ref{eq:transformer}. 
The output feature is a weighted sum of values based on the attention computed from the equation.~\ref{eq:transformer}.
In this paper, we use the encoder of the transformer with input nodes representing different objects to model the interaction between the objects. 
For more details about the transformers, we refer you to the original paper~\cite{vaswani2017attention}. 
\begin{equation}
    \texttt{Attention}(Q,K,V) = \texttt{softmax}(\frac{QK^{T}}{\sqrt{d_{k}}})V
    \label{eq:transformer}
\end{equation}

\subsection{Details of UVOS model \cite{yuan2023maximal}}

We provide implementation details about the UVOS model. In accordance with~\cite{yuan2023maximal}, we have employed Solov2~\cite{wang2020solov2} as a segmentor and STM~\cite{oh2019video} as a tracker. To familiarize the segmentor with our environment, we fine-tuned it using annotations for both environment segments and objects from YCB-Video dataset~\cite{xiang2017posecnn}.

\begin{figure}[b]
     \centering
     \includegraphics[width=0.49\textwidth]{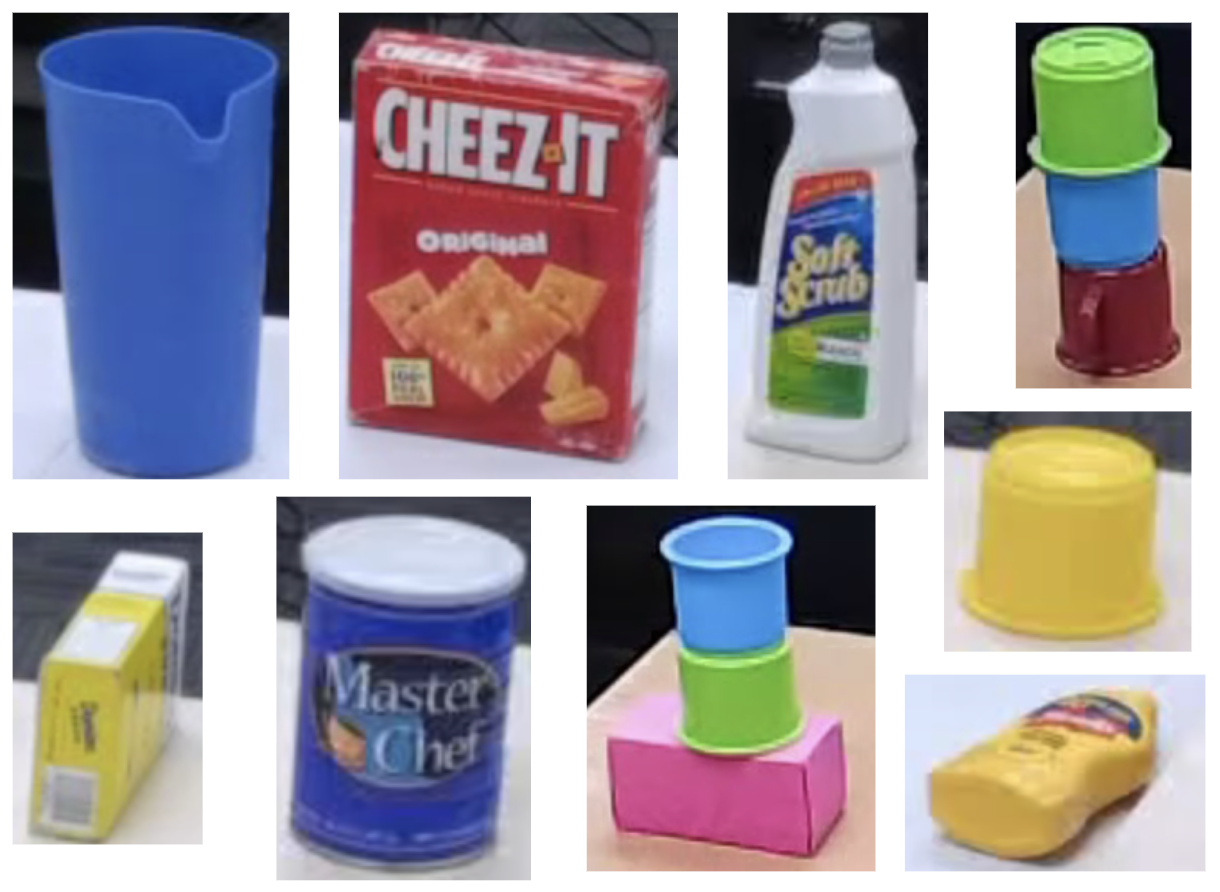}
    
     \caption{Visualization of objects in the training set for UVOS model.}
     \label{fig:labeled-objects}
\end{figure}

To be more specific, we annotated a total of $175$ images, which included objects and environments. The environment segments encompass elements such as the robot, table, and shelf. The annotated objects are visually presented in Fig.~\ref{fig:labeled-objects}, while objects labeled in the YCB-Video dataset can be found in~\cite{xiang2017posecnn}. To prevent dataset imbalance issues, we down-sampled the YCB-Video dataset before integrating it with the annotated images. As a result, our training dataset comprised $1174$ images and $7105$ object masks. Subsequently, the segmentor was fine-tuned with a learning rate of $1e-6$ over $100$ epochs. Note that the segmentor generalizes to novel objects not used during training, as shown in our real-world demonstrations.

In our tasks, key frames in the UVOS approach are designated when the robot is in its home configuration. For each key frame, the UVOS approach employs the segmentor to detect objects within the scene and utilizes the tracker to maintain consistent IDs for previously tracked objects, while those that do not match any previous object are assigned new IDs. During non-key frames, the tracker monitors all segments, relying on its object memory for tracking. 

\subsection{Skill primitives}

We define three skill primitives (push, pick-and-place, pick-and-dump) in this paper. 
We use the same push and pick-and-place skill as~\cite{Huang-icra2023-graph-relations}. 
For the details of pick-and-dump skills, we use the same grasp as the pick-and-place skill. 
After the robot grasps the object, it moves the grasped object to the dump pose. 
Then the robot rotates the joint 7 by 180 degrees to dump the contained objects. 
After the dump action, the joint 7 of the robot returns to the original value before the dump action. 

\begin{figure}
\includegraphics[width=0.98\columnwidth,clip,trim=0mm 0mm 0mm 0mm]{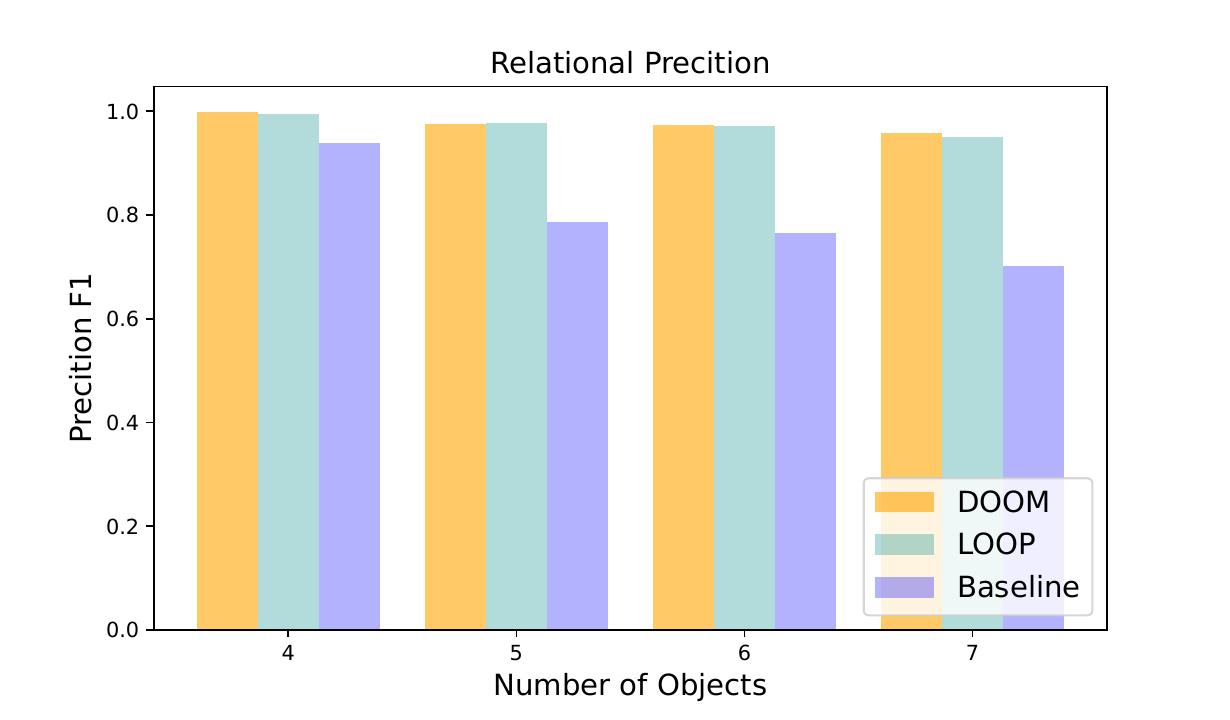} %
    \caption{Relational prediction F1 score of the different models as a function of the number of objects in the scene. }\vspace{-18pt}
    \label{fig:relational_prediction_num}
\end{figure}

\begin{figure}
\includegraphics[width=0.98\columnwidth,clip,trim=0mm 0mm 0mm 0mm]{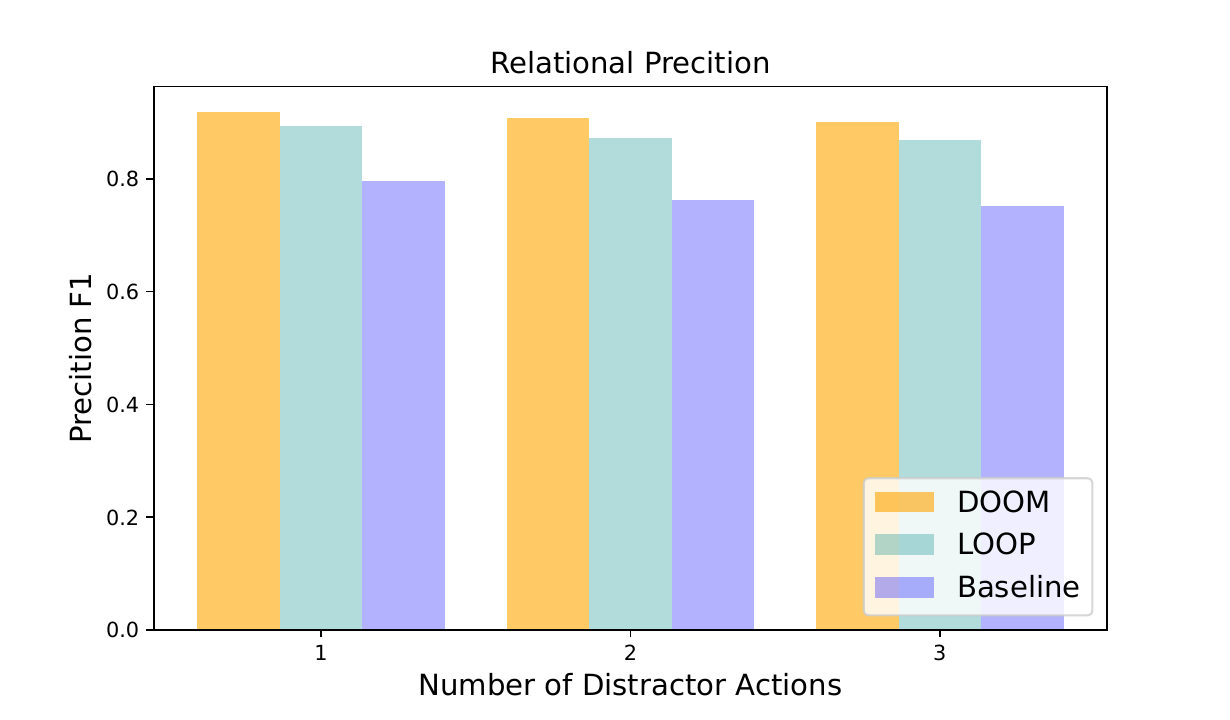} %
    \caption{Relational prediction F1 score of the different models as a function of the number of distractor actions in the scene. }\vspace{-18pt}
    \label{fig:relational_prediction_distract}
\end{figure}

\subsection{Model details}

\begin{figure*}[t]
    \centering
    \includegraphics[width=2\columnwidth]{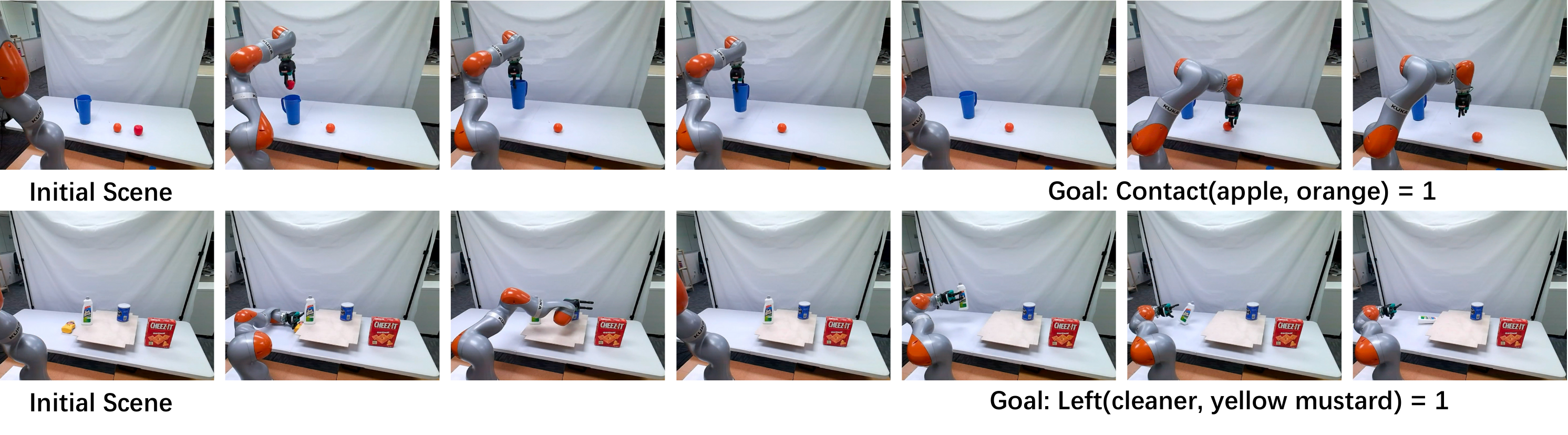}
    \caption{Two failure cases of our approaches. The first failure case is caused by a failed grasp while the second failure case is caused by unstable placement.}
\label{fig:real_failure}\vspace{-5pt}
\end{figure*}
The transformers we used in the paper consist of 2 attention layers, each layer is a 2-head self-attention block.  
The width of each attention block is 256. 
The transformer encode takes object features coming from UVOS as input and outputs the object-oriented memory. 
The dynamics model takes both the object-orientated memory and action tokens as input.
For the output of the dynamics model, we discard the action token and only keep the object tokens. 
For the decoders, the pose decoder $D_p$ is a 2-layer MLP with ReLU as an activation function. 
The relation decoder $D_r$ and environment identity decoder $D_e$ are 2-layer MLPs with Sigmoid as an activation function.  

The input for our model is partial-view point clouds. 
After the segmentation from UVOS, we use farthest point sampling to downsample each segment to 128 points. 
The PointConv model consists of 3 layers and outputs a 128-dimension feature for each segment. 
The learned positional embedding also outputs a 128-dimension feature per ID. 

For the details of our \transa{} model, $h_{type}$ is a 128 dimension feature per segment to encode whether this segment is observable at the current step. 
The $MLP$ contains 2 layers with ReLU as an activation function and outputs a 256-dimension feature. 
We train our approaches and \transa{} using the Adam optimizer with $1e-4$ as a learning rate. 

\subsection{Limitations}
We acknowledge several limitations of the proposed approach. First, we have not integrated the planner and tracker so our framework cannot achieve real-time planning. Second, while we show diverse tasks with different objects like containers and shelves, we have not included demonstrations of more compelling tasks like opening/closing drawers, which would require a robust impedance controller. Third, as shown in the problem definition, our framework assumes a known history of how the objects become occluded, limiting its ability to address occlusion without history.

\subsection{Failure case analysis}
We show two failure cases of our approaches in Fig.~\ref{fig:real_failure}. 
For the first example, the robot first puts the apple inside the pitcher and then moves the pitcher a bit. 
Then the robot receives a goal of putting the orange in contact with the apple. 
The robot plans to pick the orange but the grasp fails. 
For the second example, the robot first pushes the yellow mustard below the shelf and then pushes the coffee can. 
Then the robot picks and places the white cleaner to achieve the goal relation as \textit{left}(cleaner, yellow mustard) = 1. 
However, the placement of the cleaner is unstable and thus the goal relation is not successfully achieved. 
These two failure cases are mainly caused by low-level skill execution failures. 
The failure cases motivate future directions to implement better low-level skills and incorporate the low-level skills into our high-level planning framework. 

\subsection{Extra simulation results}
We show extra visualizations of the comparison to baseline in Fig.~\ref{fig:relational_prediction_num} and Fig.~\ref{fig:relational_prediction_distract}. 
During the comparison, we find that our approaches \pca{} and \lsa{} consistently outperform the baseline in terms of the relational prediction F1 score.

\end{document}